\documentclass[a4paper]{article}

\usepackage{INTERSPEECH2022}
\usepackage{multirow}
\usepackage[table,xcdraw]{xcolor}
\usepackage{url}

\title{Self-Supervised Speech Representations Preserve Speech Characteristics while Anonymizing Voices}
\name{Abner Hernandez$^1$, Paula Andrea P\'erez-Toro$^{1,2}$,
Juan Camilo V\'asquez-Correa$^{2,3}$, Juan Rafael Orozco-Arroyave$^{1,2}$, Andreas Maier$^1$, Seung Hee Yang$^4$}

\address{
  $^1$Pattern Recognition Lab. Friedrich-Alexander Universit{\"a}t Erlangen-N{\"u}rnberg, Germany\\
  $^2$GITA Lab. Facultad de Ingenier\'ia. Universidad de Antioquia UdeA, Medell\'in, Colombia \\
  $^3$Pratech group SAS, Medell\'in, Colombia\\
  $^4$Speech \& Language Processing Lab, Friedrich-Alexander Universit{\"a}t Erlangen-N{\"u}rnberg, Germany}
\email{abner.hernandez@fau.de, paula.andrea.perez@fau.de, jcamilo.vasquez@udea.edu.co, rafael.orozco@udea.edu.co, andreas.maier@fau.de, seung.hee.yang@fau.de}

\begin{document}

\maketitle
\begin{abstract}
  Collecting speech data is an important step in training speech recognition systems and other speech-based machine learning models. However, the issue of privacy protection is an increasing concern that must be addressed. The current study investigates the use of voice conversion as a method for anonymizing voices. In particular, we train several voice conversion models using self-supervised speech representations including Wav2Vec2.0,  Hubert and UniSpeech. Converted voices retain a low word error rate within 1\% of the original voice. 
  Equal error rate increases from 1.52\% to 46.24\% on the LibriSpeech test set and from 3.75\% to 45.84\% on speakers from the VCTK corpus which signifies degraded performance on speaker verification. 
  Lastly, we conduct experiments on dysarthric speech data to show that speech features relevant to articulation, prosody, phonation and phonology can be extracted from anonymized voices for discriminating between healthy and pathological speech. 

\end{abstract}

\section{Introduction}
Speech-based technology such as virtual assistants is increasingly common use in everyday life. While these technologies can make life easier, there is a risk of users' voices being stored or sent to third-party organizations. The European Union's General Data Protection Regulation notes that restrictions and regulations on the use of user speech data are essential~\cite{nautsch19}.

However, improving speech recognition performance requires training machine learning systems on a large amount of user speech data in natural settings. The need for collecting speech data that respects users' privacy has led to voice privacy research. Voice anonymization involves removing personally identifiable information from a speech signal to remove the speaker's identity. The same aim is required when processing pathological speech, where
privacy is also necessary to be preserved prior to transmit recordings for further analyses, diagnoses or screenings.
The VoicePrivacy 2020 Challenge~\cite{tomashenko2022} was the first challenge to focus on voice anonymization for speech technology. The performance of anonymization is evaluated using Equal Error Rate (EER) to determine how well voices are anonymized and Word Error Rate (WER) to measure how well anonymized speech signal retains information for automatic speech recognition (ASR). 

Various approaches to voice anonymization have been explored. Some methods include adding noise to the speech signal~\cite{hashimoto2016}, transforming the spectral envelope via McAdams coefficient~\cite{patino21}, adversarial training~\cite{perero2022,srivastava19}, and more recently voice conversion~\cite{Yoo2022,gaznepoglu2021,qian2019}. Voice conversion is the process of transforming the voice of a source speaker to a target speaker while maintaining the linguistic content but not the source speaker content. 

In the current study, we train voice conversion models with self-supervised pre-trained representations as input to models. Self-supervised representation learning involves training with a large amount of unlabelled data and learning a speech representation that can be used for downstream tasks. The popular Wav2Vec2.0~\cite{Baevski2020} model takes a raw waveform as input to convolutional layers. The output is quantized, masked and then fed to a transformer-based context network. The objective of pre-training Wav2Vec2.0 is to use a contrastive loss function to identify the true quantized representation of masked segments given a set of candidate representations. The final quantized representations can then be used on a downstream task in replacement of traditional features such as Mel-frequency cepstral coefficients (MFCC) or Filterbanks (Fbanks).
The quality of anonymization is then evaluated on the LibriSpeech-test~\cite{panayotov2015}, and VCTK~\cite{yamagishi2019} datasets. Experimental results suggest that voice conversion models trained on self-supervised representations reduce the performance of speaker verification while maintaining a WER similar to unaltered speech. Results varied depending on the model and the gender of the source and target speaker. In general, converting all male voices to a female source speaker and all female voices to a male source speaker led to the best anonymization results, while converting all voices to a female target led to the lowest WER.

Furthermore, we investigate whether the anonymized voices retain linguistic and acoustic information useful for classifying pathological speech. We train random forest classifiers on the UASpeech corpus~\cite{kim08} using speech features related to articulation~\cite{orozco2018}, prosody~\cite{orozco2020current}, phonation~\cite{vasquez2021modeling} and phonology~\cite{vasquezcorrea19}. Results show that classifiers trained on articulatory and phonological features are minimally affected by anonymization compared to prosody and phonation features. However, the results varied depending on the specific speech representation used for training.

The rest of the article is organized as follows: Section 2 will give a brief overview of related voice anonymization studies, specifically, ones using voice conversion. In Section 3 we describe our methodology of training voice conversion models with self-supervised pre-training and the classification of pathological speech. We also go over the results of all experiments. A conclusion and future directions are discussed in Section 4.

\section{Related Work}
The use of voice conversion for anonymization has been explored in several studies. A classical approach is to use the vocal tract length normalization (VTLN)~\cite{sundermann2003} algorithm which modifies the spectrum of each frame using frequency warping. This approach was used in ~\cite{qian2019} within a 6-step anonymization process where a speech signal undergoes the following: pitch marking, frame segmentation, fast Fourier transform on the frequency domain, VTLN, inverse fast Fourier transform, and PSOLA (Pitch-Synchronous Overlap-Add). The speaker identification accuracy was reduced by 83.7\% but also a 19.1\% drop in speech recognition accuracy.

The best model from the VoicePrivacy 2020 Challenge~\cite{tomashenko2022} was a semi-adversarial training-based method~\cite{champion2020}, which increased the EER from 3.29\% to 53.37\%. However, when evaluating speech recognition performances on the LibriSpeech-test set, the WER increases from the original 4.1\% to 8.5\%. The model which best maintained WER for the Librispeech test-set was a signal-based processing system~\cite{dubagunta2020} that modifies the formants, fundamental frequency and speech rate. While the WER was 5.8\%, the EER only reached 28.19\%. Results from all submitted models show that anonymization always degrades speech recognition performance, with a relative WER increase between 40-217\% for the LibriSpeech-test dataset, and between 14-120\% on the VCTK-test set. 

Table~\ref{tab:vp-chall} shows the results from the top 5 models in terms of high EER. All models utilize an x-vector based anonymization method. The singular value decomposition of an utterance-level speaker x-vector was used for A1, while variability-driven decomposition of x-vectors was used in A2. S1 and S2 models use domain-adversarial training to generate x-vectors. Lastly, O1 and O1c1 systems use Gaussian mixture models to sample x-vectors in a PCA-reduced space.

\begin{table}[th]
\caption{Voice anonymization results from the VoicePrivacy2020 challenge~\cite{tomashenko2022}.}
\label{tab:vp-chall}
\centering
\begin{tabular}{cccc}
\multicolumn{1}{l}{Dataset} & \multicolumn{1}{l}{Model}                                                     & \multicolumn{1}{l}{EER (\%)}                                                          & \multicolumn{1}{l}{WER (\%)}                                                    \\ \hline
LibriSpeech-test            & \begin{tabular}[c]{@{}c@{}}Original\\ M1c1\\ M1\\ K2\\ A2\\ A1\end{tabular}   & \begin{tabular}[c]{@{}c@{}}3.29*\\ 53.37\\ 52.23\\ 52.38\\ 51.87\\ 51.11\end{tabular} & \begin{tabular}[c]{@{}c@{}}4.1\\ 8.5\\ 8.4\\ 13.2\\ 6.8\\ 13.2\end{tabular}     \\ \hline
VCTK-test                   & \begin{tabular}[c]{@{}c@{}}Original\\ O1c1\\ O1\\ S2\\ S2c1\\ S1\end{tabular} & \begin{tabular}[c]{@{}c@{}}3.29\\ 39.79\\ 39.34\\ 39.16\\ 38.26\\ 36.92\end{tabular}  & \begin{tabular}[c]{@{}c@{}}12.8\\ 15.6\\ 15.6\\ 15.2\\ 15.2\\ 15.5\end{tabular} \\ \hline
\end{tabular}

\textit{*EER is calculated with the average of both LibriSpeech and VCTK datasets.}
\end{table}


\section{Experiments and Results}
The first part of this section presents the experiments and results
obtained when evaluating different voice conversion methods and their impact
on speaker verification. The second part evaluates to which extent
voice conversion retains linguistic and acoustic information relevant to
discriminate between dysarthric and healthy speech signals.
\subsection{Voice Conversion Experiment}
\begin{figure}[htb]
    \centering
    \centerline{\includegraphics[width=5.0cm]{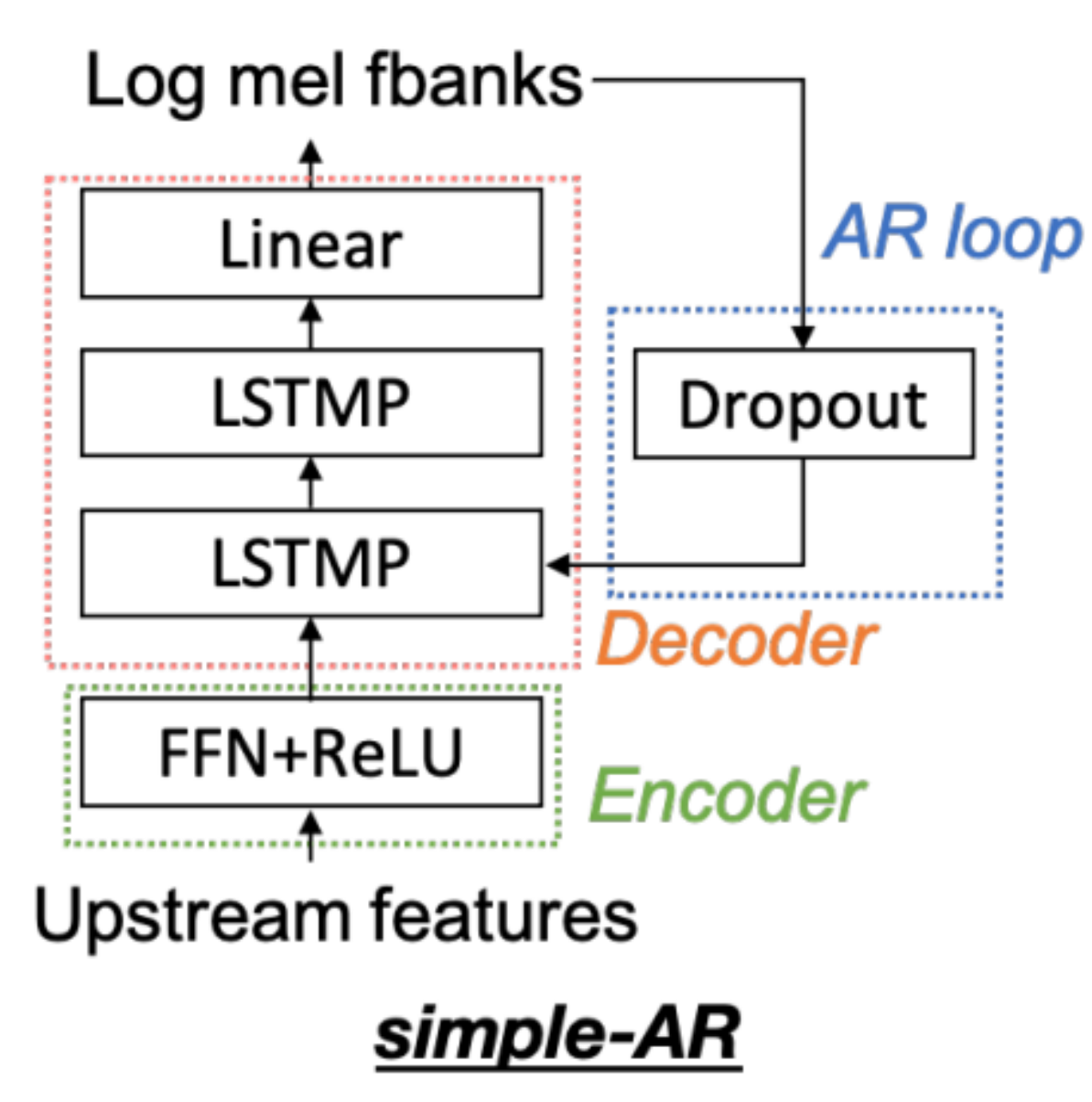}}
    \caption{Voice conversion model where Upstream features refer to self-supervised representations~\cite{huang2021}.}
    \label{fig:simple-ar}%

\end{figure}

The S3PRL-VC toolkit~\cite{huang2021} is used to train an encoder-decoder based voice converter. As seen in Figure~\ref{fig:simple-ar}, the model contains a single layer feedforward network, two long short-term memory layers with projection, and a linear projection layer. The target acoustic feature is the Mel-spectrogram. An autoregressive model is included as it has been shown to be useful in speech synthesis~\cite{wang2017}.

Multiple voice conversion models are trained with the following 7 self-supervised representations: Wav2Vec~\cite{schneider19}, Wav2Vev2.0~\cite{Baevski2020}, Hubert~\cite{Hsu2021}, WavLM~\cite{chen2021}, Unispeech-SAT~\cite{chen2021unispeech}, XLSR-53~\cite{conneau21}, and XLSR-128~\cite{babu2021}. XLSR-53 is a cross-lingual Wav2Vec2.0 model trained on 53 different languages, while XLSR-128 is trained on 128 languages. For each pre-trained model, a voice converter is trained on a female and male target speaker, leading to 14 models. The models are trained on the Voice Conversion Challenge 2020 database~\cite{zhao2020}. The log-Mel spectrograms are converted to speech samples via the Griffin-Lim algorithm. 

When evaluating the LibriSpeech data we use the clean test set which contains 20 male speakers and 20 female speakers, see Table~\ref{tab:eval-data}. In total, there are 2,620 utterances. As there is no official test set for the VCTK data, we randomly select 5 male and 5 female speakers with American or Canadian dialects. There are 3,989 utterances in total.

\begin{table}[th]
\caption{Evaluation data information for LibriSpeech and VCTK datasets.}
\label{tab:eval-data}
\centering
\begin{tabular}{c c c c}
Corpus & Females & Males & Utterances\\
\hline\
LibriSpeech & 20 & 20 & 2,620\\
VCTK  & 5 & 5 & 3,989\\
\hline
\end{tabular}
\end{table}

\subsubsection{Evaluation}
Anonymization is evaluated using EER. Matching and non-matching pairs are created and a pre-trained speaker verification model is used to determine whether a pair is from the same speaker or not. We used an attention-based TDNN speaker verification model~\cite{desplanques20} which achieves a 0.69\% EER on the VoxCeleb test set.
With the LibriSpeech data, for each utterance, we randomly select 5 audio files from the same speaker and 5 audio files from other speakers. This leads to a total of 26,200 trial pairs. For the VCTK data, we randomly select 3 utterances from the same speaker and 3 from other speakers. Leading to a total of 31,912 trial pairs. A Wav2Vec2.0 model pre-trained and fine-tuned on 60,000 hours of data and using a self-training objective~\cite{xu2021self} was used to measure WER. This model achieves a 1.9\% WER on the test-clean set of LibriSpeech.

\subsubsection{Voice Conversion Results}
Results from Table~\ref{tab:eer-res} show EER values from all models in both female and male trained sessions. The EER for the original audio files is 1.53\% for the LibriSpeech data and 1.39\% for the VCTK data. In both cases, converting voices with a male target speaker and using XSLR-53 speech representations reaches the highest EER at 41.18\% and 42.19\% for the LibriSpeech and VCTK data respectively. 

\begin{table}[th]
\caption{EER (\%) results for the LibriSpeech and VCTK datasets. Results from converting all voices to a male target speaker are in parenthesis, while outside the parenthesis refers to converting all voices to a female target speaker.}
\label{tab:eer-res}
\centering
\begin{tabular}{c c c}
Model & LibriSpeech (male) & VCTK (male)\\
\hline\
Original & 1.53 & 1.39\\
Wav2Vec & 34.77 (33.69) & 28.75 (35.33)\\
Wav2Vec2.0 & 32.36 (31.05) & 24.07 (34.01)\\
Hubert & 27.77 (27.8) & 21.82 (32.19)\\
WavLM & 29.5 (29.4) & 22.63 (34.23)\\
UniSpeech-SAT & 29.15 (29.03) & 23.12 (33.94)\\
XLSR-53 & 38.53 (41.18) & 30.77 (42.19)\\
XLSR-128 & 31.7 (28.24) & 23.18 (34.24)\\
\hline
\end{tabular}
\end{table}

\begin{figure}[htb]
    \centering
    \centerline{\includegraphics[width=7.0cm]{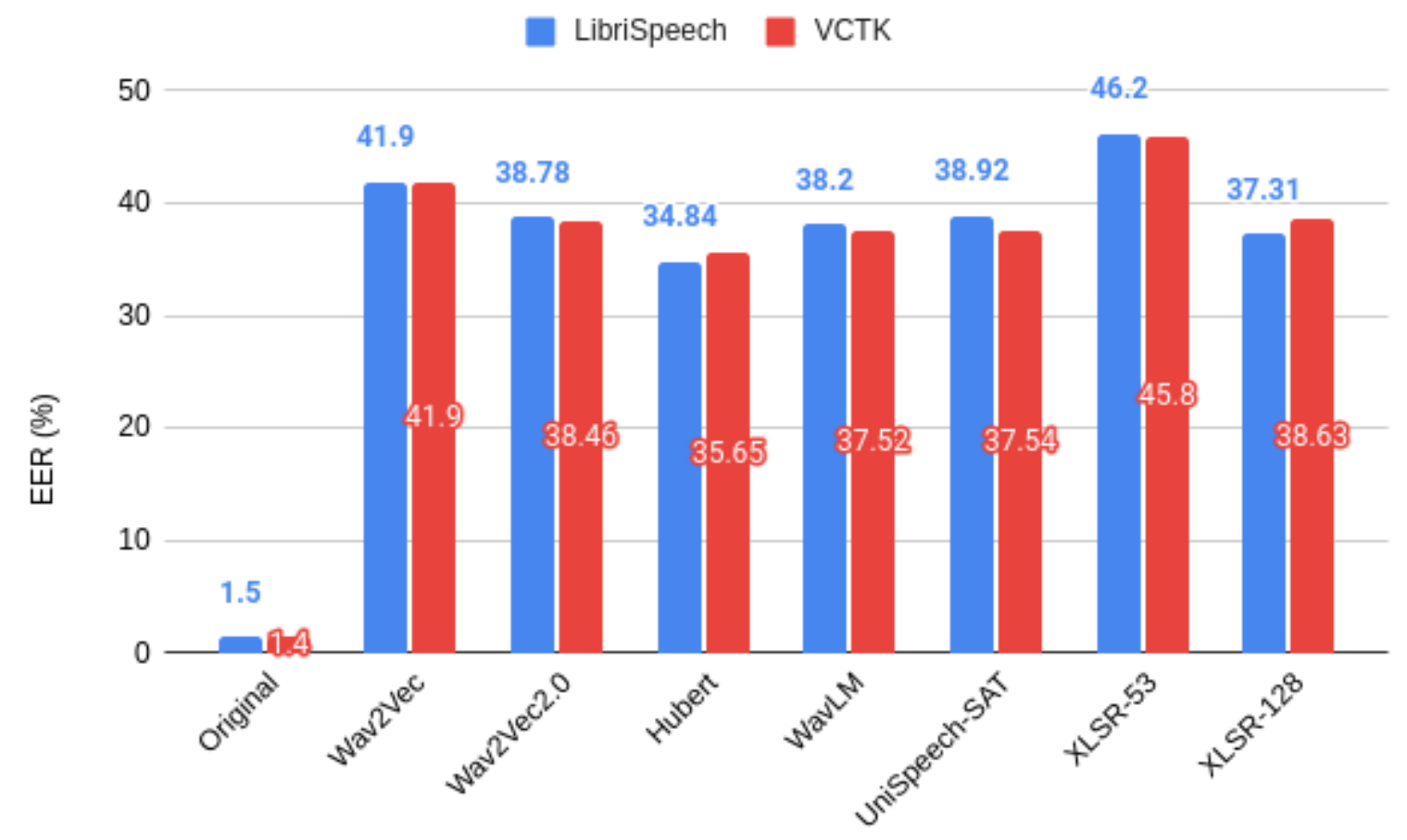}}
    \caption{EER results when converting source male voices to female target voices and vice-versa.}
    \label{fig:mixed-eer}%

\end{figure}

We also looked at the effect of converting voices based on the opposite gender. From Table~\ref{fig:mixed-eer} we see that in all cases, converting female to male voices and vice-versa results in high EERs. In both datasets, the XLSR-53 model increases the EER to 46.24\% and 45.84\% for the LibriSpeech and VCTK data respectively.

\begin{figure}[htb]
    \centering
    \centerline{\includegraphics[width=7.0cm]{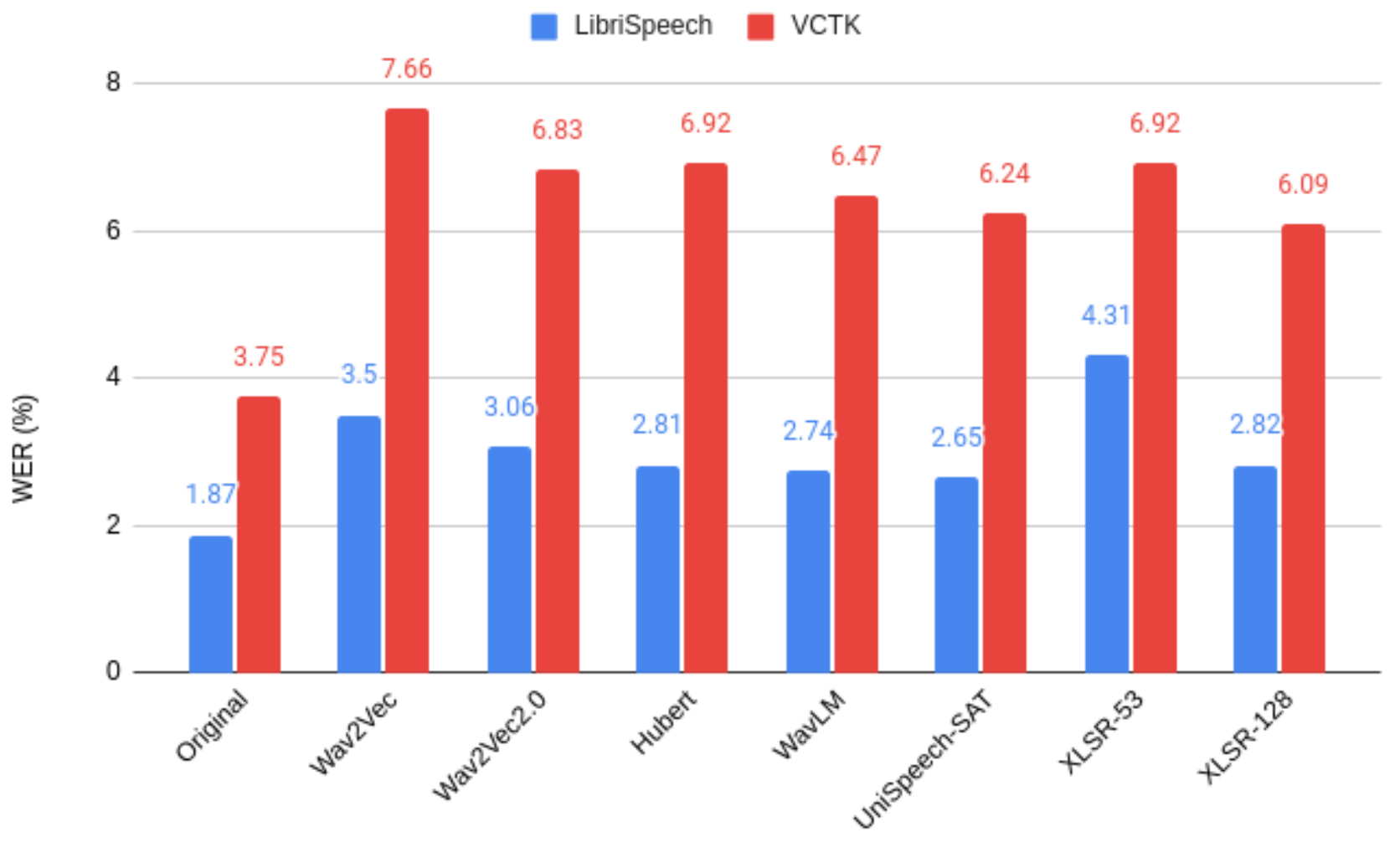}}
    \caption{WER results when converting source male voices to female target voices and vice-versa.}
    \label{fig:mixed-wer}%

\end{figure}

Results of WER evaluation can be seen in Table~\ref{tab:wer-res}, the models which had a high EER tend to also have a high WER. This shows that there is a trade-off between anonymity and preserving speech quality. Furthermore, WER is less affected by anonymization with the LibriSpeech data compared to the VCTK data. WER's of 1.87\% and 3.75\% is obtained with the non-anonymized voices for the LibriSpeech and VCTK data respectively. The lowest WER (2.46\%) for LibriSpeech came from the UniSpeech-SAT trained on a female target speaker. While the XLSR-128 trained on a female target speaker lead to the lowest WER for the VCTK data. As seen in Figure~\ref{fig:mixed-wer} WER results when converting voices based on gender does not produce better results than simply converting all speech to a female voice.

\begin{table}[th]
\caption{WER (\%) results for the LibriSpeech and VCTK datasets. Results from converting all voices to a male target speaker are in parenthesis, while outside the parenthesis refers to converting all voices to a female target speaker.}
\label{tab:wer-res}
\centering
\begin{tabular}{c c c}
Model & LibriSpeech (male) & VCTK (male)\\
\hline\
Original & 1.87 & 3.75\\
Wav2Vec & 3.17 (3.67) & 7.36 (7.46)\\
Wav2Vec2.0 & 2.74 (2.92) & 6.39 (6.27)\\
Hubert & 2.61 (2.75) & 6.14 (6.39)\\
WavLM & 2.56 (2.7) & 5.92 (6.15)\\
UniSpeech-SAT & 2.46 (2.69) & 5.82 (5.88)\\
XLSR-53 & 3.69 (3.99) & 8.11 (9.05)\\
XLSR-128 & 2.62 (2.69) & 5.65 (5.79)\\
\hline
\end{tabular}
\end{table}

\vspace{-0.4cm}
\subsection{Pathological Speech Experiment}
For the experiments on dysarthric speech signals using the UASpeech data, we used the Wav2Vec, XLSR-53, and UniSpeech-SAT models as these three yielded the best performance in terms of WER and EER with both LibriSpeech and VCTK datasets. Only speakers with dysarthria have their voice converted. We randomly select 7 speakers with mid to severe dysarthria. 2,200 trial pairs of matching and non-matching audio pairs are created. The trial pairs contain the same audio that will be used when evaluating classification accuracy. The original EER of is quite high compared to healthy speech but increases from 16.45\% to 43.46\% when anonymizing with XLSR features.

\begin{table}[th]
\caption{EER results for the UASpeech test set.}
\label{tab:pitch-res}
\centering
\begin{tabular}{c c c c}
Model & EER\\
\hline\
Original & 16.45\\
XLSR  & \textbf{43.46}\\
Wav2Vec  & 34.91\\
UniSpeech-SAT  & 34.18\\
\hline
\end{tabular}
\end{table}

\subsubsection{Feature Extraction}

We extracted four different feature sets commonly used to assess pathological speech to validate whether the disease condition persists after the anonymization procedure. Features are extracted using the DisVoice toolkit\footnote{\url{https://github.com/jcvasquezc/DisVoice}}, and include articulation, prosody, phonation and phonology.

Articulation features model the ability of the patients to control the articulatory muscles. The features considered the energy content in the transition between voiced and unvoiced segments~\cite{orozco2018}. The transition segments are detected based on the computation of the fundamental frequency ($F\raisebox{-.4ex}{\scriptsize 0}$). Once the border between unvoiced and voiced segments is detected, 40\,ms of the signal are taken to the left and the right. The spectrum of the transition segments is distributed into 22 critical bands according to the Bark scale, and the Bark-band energies are calculated. 12 MFCCs and their first two derivatives are also computed in the transitions to complete the feature set.

Prosody features are designed to model monotonicity, monoloudness, and speech rate disturbances in the patients. The total feature set is divided into three groups to model the $F\raisebox{-.4ex}{\scriptsize 0}$ contour (30 features), the energy contour (48 features), and the duration and speech rate (25 features). A complete description of the feature set can be found in~\cite{orozco2020current}.

Phonation features aim to model abnormal patterns in the vocal fold vibration. The feature set includes descriptors computed for 40\,ms frames of speech from voiced segments, including jitter, shimmer, amplitude perturbation quotient, pitch perturbation quotient, the first and second derivatives of the $F\raisebox{-.4ex}{\scriptsize 0}$, and the log-energy~\cite{vasquez2021modeling}.

Finally, phonology features are represented by a vector with interpretable information about the placement and manner of articulation. The different phonemes that appear in a certain language are grouped into 18 phonological posteriors~\cite{vasquezcorrea19}. The phonological posteriors were computed with an array of recurrent neural networks to estimate the probability of occurrence of a specific phonological class. Then, the posterior probabilities are projected into phonological log-likelihood ratio (PLLRs) features~\cite{diez2014projection} to overcome the non-Gaussian nature of phonological posteriors and to avoid bounding effects, which is better to exploit different classification methods.

\subsubsection{Classification Procedure}
We use Scikit-learn~\cite{scikit-learn} to train random forest classifiers on 9,737 utterances from 21 speakers and evaluated on 3,255 utterances of a separate test set. All classifiers have 100 decision trees with a maximum depth of 20. To measure the quality of a split we use entropy for the information gain. We train models on the original audio, XLSR-anonymized audio, Wav2Vec-anonymized audio, and UniSpeech-anonymized audio. Classifiers are trained and tested with the same audio modifications (e.g. train with XLSR-anonymized audio, tested with XLSR-anonymized audio). In each case, we separately train on features relevant for articulation, prosody, phonation and phonology.

\subsubsection{Classification Results}
\begin{figure}[htb]
    \centering
    \centerline{\includegraphics[width=7.0cm]{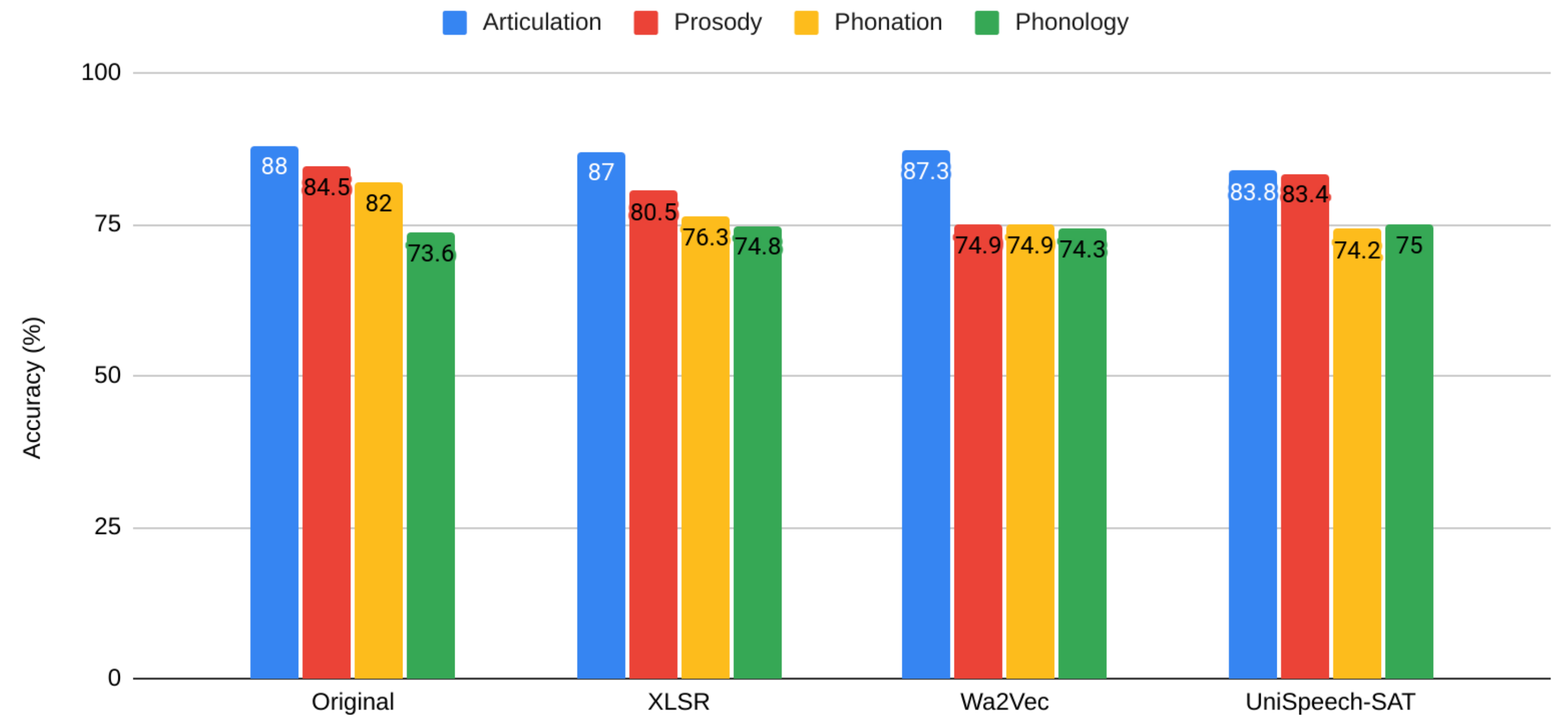}}
    \caption{Classification results when extracting speech features from original and anonymized voices.}
    \label{fig:ua-results}%

\end{figure}

Results from Figure~\ref{fig:ua-results} show that in all cases, articulatory features lead to the highest accuracy. Models trained on the original audio reached an accuracy of 88\% while scores of 87\%, 87.3\%, and 83.8\% were obtained by the XLSR, Wav2Vec, and UniSpeech-SAT models respectively. Classifiers trained on prosodic features are more affected by anonymization. The original accuracy is 84.5\% but drops to 80.5\% and 74.9\% for XLSR and Wav2Vec models. The UniSpeech-SAT trained model is less affected with an accuracy of 83.4\%. Anonymization also negatively affects classifiers trained on phonation-based features. The original accuracy of 82\% drops to 76.3\%, 74.9\%, and 74.2\% with XSLR, Wav2Vec, and UniSpeech-SAT models respectively. Lastly, the accuracy of classifiers trained on phonology features with original audio samples is 73.6\% but increases to 74-75\% with models trained on anonymized voices. As seen in Table~\ref{tab:avg-res} the classifier trained on XLSR features was the closest in terms of F1-score towards of F1-score, with a difference of 2.6\% compared to the original score.

\begin{table}[th]
\caption{Classification performance when averaging the results from all speech features.}
\label{tab:avg-res}
\centering
\begin{tabular}{c c c c}
Model & Precision & Recall & F1-Score\\
\hline
Original & 88.1 & 79.7 & 83.7\\
\hline
XLSR & \textbf{88.1} & 75.1 & \textbf{81.1}\\
Wav2Vec & 83.9 & \textbf{76.5} & 80\\
UniSpeech-SAT & 85.9 & 76.0 & 80.6\\
\hline
\end{tabular}
\end{table}

\section{Conclusions}
In this study, we show that self-supervised representations can be trained for voice conversion and used as a method for voice anonymization. Specifically, we look at Wav2Vec, Wav2Vec2.0, Hubert, WavLM, UniSpeech-SAT, XLSR-53 and XLSR-128 pre-trained models. The XLSR-53 model was the best at increasing EER. This occurs when all source speakers are converted to a target speaker of the opposite gender. Otherwise converting all source speakers to a male target speaker leads to higher EER than a female target speaker. An increase from 1.53\% to 46.2\% and an increase from 1.39\% to 45.8\% for the LibriSpeech and VCTK datasets respectively. 
However, retaining the utility of anonymized voice was best maintained by the UniSpeech-SAT and XLSR-128 models. A WER increase of 0.59\% was seen in the Librispeech data, while an increase of 1.9\% was seen for the VCTK data. Lastly, we show that anonymized pathological speech retains some acoustic information that distinguishes healthy speech from dysarthric speech. In particular, articulatory and phonological features showed minimal differences in classification accuracy compared to non-anonymized voices. A future plan is to conduct a deeper analysis of the phonemic distribution in anonymized voices and compare how well preserved they are in comparison to non-anonymized speech.

\bibliographystyle{IEEEtran}

\bibliography{mybib}

\end{document}